\documentclass[11pt,a4paper]{article}
\usepackage[hyperref]{naaclhlt2018}
\usepackage{times}
\usepackage{latexsym}
\usepackage{array}
\usepackage{xcolor} 
\usepackage{verbatim} 
\usepackage{subcaption}
\usepackage{booktabs}
\usepackage[draft]{pgf}
\usepackage{url}
\usepackage{amsmath}
\usepackage{amssymb}

\aclfinalcopy % Uncomment this line for the final submission
\def\aclpaperid{6} %  Enter the acl Paper ID here

\newcommand*{\affaddr}[1]{#1} % No op here. Customize it for different styles.

\newcommand*{\email}[1]{\texttt{#1}}

\title{The Fine Line between Linguistic Generalization and Failure in Seq2Seq-Attention Models }

\author{%
Noah Weber\thanks{*These authors contributed equally to this work.}, Leena Shekhar\footnotemark[1], Niranjan Balasubramanian \\
\affaddr{Stony Brook University, NY}\\
\email{\{nwweber, lshekhar, niranjan\}@cs.stonybrook.edu}\\
}

\date{\today}

\begin{document}
\maketitle
\begin{abstract}
Seq2Seq based neural architectures have become the go-to architecture to apply to 
sequence to sequence language tasks. 
Despite their excellent performance on these tasks, recent work has noted that these models 
usually do not fully capture the linguistic structure required to generalize beyond the dense sections of the data distribution \cite{ettinger2017towards}, and as such, are likely to fail on samples from the tail end of the distribution (such as inputs that are noisy \citep{belkinovnmtbreak} or of different lengths \citep{bentivoglinmtlength}).
In this paper, we look at a model's ability to generalize on a simple symbol rewriting task with a clearly defined structure. We find that the model's ability to generalize this structure beyond the training distribution depends greatly on the chosen random seed, even when performance on the standard test set remains the same. 
%finding
This suggests that a model's ability to capture generalizable structure is highly sensitive. Moreover, this sensitivity may not be apparent when evaluating it on standard test sets. 
\end{abstract}

\section{Introduction}

It is well known that language has certain structural properties which allows natural language speakers to make ``infinite use of finite means" \cite{chomsky1965}.  This structure allows us to generalize beyond the typical machine learning definition of generalization \cite{valiant1984} (which considers performance on the distribution that generated the training set), permitting the understanding of any utterance sharing the same structure, regardless of probability. For example, sentences of length 100 typically do not appear in natural text or speech (our personal 'training set'), but can be understood regardless due to their structure.
We refer to this notion as \textit{linguistic} generalization \footnote{From here on, mentions of generalization refer to the linguistic kind.}.

Many problems in NLP are treated as sequence to sequence tasks with solutions built on seq2seq-attention based models. While these models perform very well on standard datasets and also appear to capture some linguistic structure~\cite{linzen2018poverty, belkinovmorph, linzenDG16}, they also can be quite brittle, typically breaking on uncharacteristic inputs~\cite{lake2018still, belkinovnmtbreak}, indicating that the extent of linguistic generalization these models 
achieve is still somewhat lacking.

Due to the high capacity of these models, it is not unreasonable to expect them to learn \textit{some} structure from the data. However, learning structure is not a sufficient condition to achieving linguistic generalization. If this structure is to be usable on data outside the training distribution, the model must learn the structure \textit{without} additionally learning (overfitting on) patterns specific to the training data. One may hope, given the right 
hyperparameter configuration and regularization, that a model converges to a solution that captures the reusable structure without overfitting too much on the training set. While this solution exists in theory, in practice, it may be difficult to find.

In this work, we look at the feasibility of training and tuning seq2seq-attention models towards a solution that generalizes in this linguistic sense. In particular, we train models on a symbol replacement task with a well defined generalizable structure. The task is simple enough that all models achieve near perfect accuracy on the standard test set, i.e., where the inputs are drawn from the same distribution as that of the training set.
%i.e., where the inputs are drawn from the same distribution that the training set is drawn from.
We then test these models for linguistic generalization by creating test sets of uncharacteristic inputs, i.e., inputs that are not typical in the training distribution but still solvable given that the generalizable structure was learned.
%even
Our results indicate that generalization is highly sensitive\footnote{The sensitivity of generalization is also hinted at in \citet{linzen2018poverty} who additionally note performance variations across initializations}; such that even changes in the random seed can drastically affect the ability to generalize. This dependence on an element that is not (or ideally \textit{should not be}) a hyperparameter suggests that the line between generalization and failure is quite fine, and may not be feasible to reach simply by hyperparameter tuning alone. 

\section{Generalization in a Symbol Rewriting Task}
%pick up
Real world NLP tasks are complex, and as such, it can be difficult to precisely define what a model should and should not learn during training. As done in previous work \cite{lake2018still,rodriguez1998recurrent}, we ease analysis by looking at a simple formal task.
The task is set up to mimic (albeit, in an oversimplified manner) the input-output symbol
alignments and local syntactic properties that models must learn in many natural language tasks, such as translation, tagging and summarization. 

The task is defined over sequences of symbols, $\{x_1,...x_n | x_i \in X\}$, where $X$ is the input alphabet. Each symbol $x \in X$ is uniquely associated with its own 
output alphabet $Y_x$. Output is created by 
taking each individual symbol $x_i$ in the sequence and rewriting it as any sequence of $k$ symbols from 
$Y_{x_i}$. To do the task, the model must learn alignments between the input and output symbols, and preserve the simple local syntactic conditions (every group of $k$ symbols must come from the same input alphabet $Y_x$). 

As an example, let $X = \{A,B\}$, $Y_A = \{\text{$a_{1}, a_{2}$, $a_3$}\}$, $Y_B = \{\text{$b_1, b_2,$ $b_3$}\}$, and $k=3$. Each $a_i$ and $b_i$ has 2 possible values, $a_{i1}$ or $a_{i2}$ and $b_{i1}$ or $b_{i2}$ respectively. Thus, mapping an input symbol to 48 ($8 * 3!$) possible permutations. A possible valid output for the input $AB$ is $a_{21}a_{32}a_{11}b_{32}b_{11}b_{22}$. Note that such valid strings are selected at random when generating the dataset. We allow this stochasticity in the outputs in order to prevent the model from resorting to pure memorization. For our task, $\vert X\vert=40$ and each $x_i$ has a corresponding output alphabet $Y_{x_i}$ of size 16.

\section{Model and Training Details}
The models we use are single layer, unidirectional, seq2seq LSTMs \citep{hochreiter1997} with bilinear attention \citep{luongPM15} and trained with vanilla SGD. To determine the epoch to stop training at, we create a validation set of 2000 samples with the same characteristics as the training set, i.e., of length 5-10 with no repeated symbols. 
Training is stopped once accuracy\footnote{We compute accuracy as $\frac{\text{\# times the model produced a valid output}}{\text{\# samples}}$.} on the validation set either decreases or remains unchanged. 
The size of the hidden state and embeddings were chosen such that they were as small 
as possible without reducing validation accuracy, giving a size of 32.

Tuning hyperparameters is often done on a validation set drawn from the same distribution as the training set (as opposed to validating on data from a different distribution) which motivated our initial decision to use a validation set of characteristic inputs to decide the epoch to stop at. However, we noticed only small variation in the validation performance upon using different learning rates and dropout probabilities (where dropout was applied to the input and output layers). In order to fine tune these parameters to avoid extreme overfitting, we created another validation set  consisting 
of 5000 samples of "uncharacteristic" inputs, i.e., inputs with repeated symbols and varying from length 3-12. These two hyperparameter values were set to 0.125 and 0.1, respectively, according to the performance on this validation set, averaged across a set of randomly chosen random seeds. All other hyperparameters were also decided using this validation set. Further training details are listed in Table~\ref{tab:model-params}.

% this table has model training details
\begin{table}%[t]
\centering
\small
\begin{tabular}{l | r}
\toprule
LSTM Layers & 1 \\
\hline
WE/LSTM size & 32 \\
\hline
Attention & Bilinear \\
\hline
Batch size & 64 \\
\hline
Optimizer & SGD \\
\hline
LR & 0.125 \\
\hline
Max gradient norm & 5 \\
\hline
Dropout & 0.1 \\
\hline
Initialization & Uniform(-0.1, 0.1)\\
\bottomrule
\end{tabular}
\caption{
\label{tab:model-params}
Model details.}
\end{table}

\section{Experiments}
To generalize to any input sequence, a model must: \textbf{(1)} learn the generalizable structure - the alignments between input and output alphabets, and \textbf{(2)} \textit{not} learn any dependencies among input symbols or sequence length. To test the extent to which (2) is met, we train seq2seq-attention models with 100,000 randomly generated samples with inputs uniformly generated with lengths 5-10 and no input symbol appearing more than once in a single sample. If the model learned alignments without picking up other dependencies among input symbols or input lengths then the resulting model should have little problem in handling inputs with repeated symbols or different lengths, despite never seeing such strings.
% To test the extent to which \textbf{(2)} is met, we train\footnote{A detailed account of model training, regularization, and tuning is provided in the supplementary material.} seq2seq-attention models with 100,000 randomly generated samples (with inputs uniformly generated with lengths 5-10), with the catch being that no input symbol appears more than once in a single sample. If the model learned alignments without picking up other dependencies among input symbols (or input length) then the resulting model should have little problem in handling inputs with repeated symbols (or different length), despite never seeing such a string.
\begin{figure}%[t]
    \centering
        \includegraphics[width=1\columnwidth]{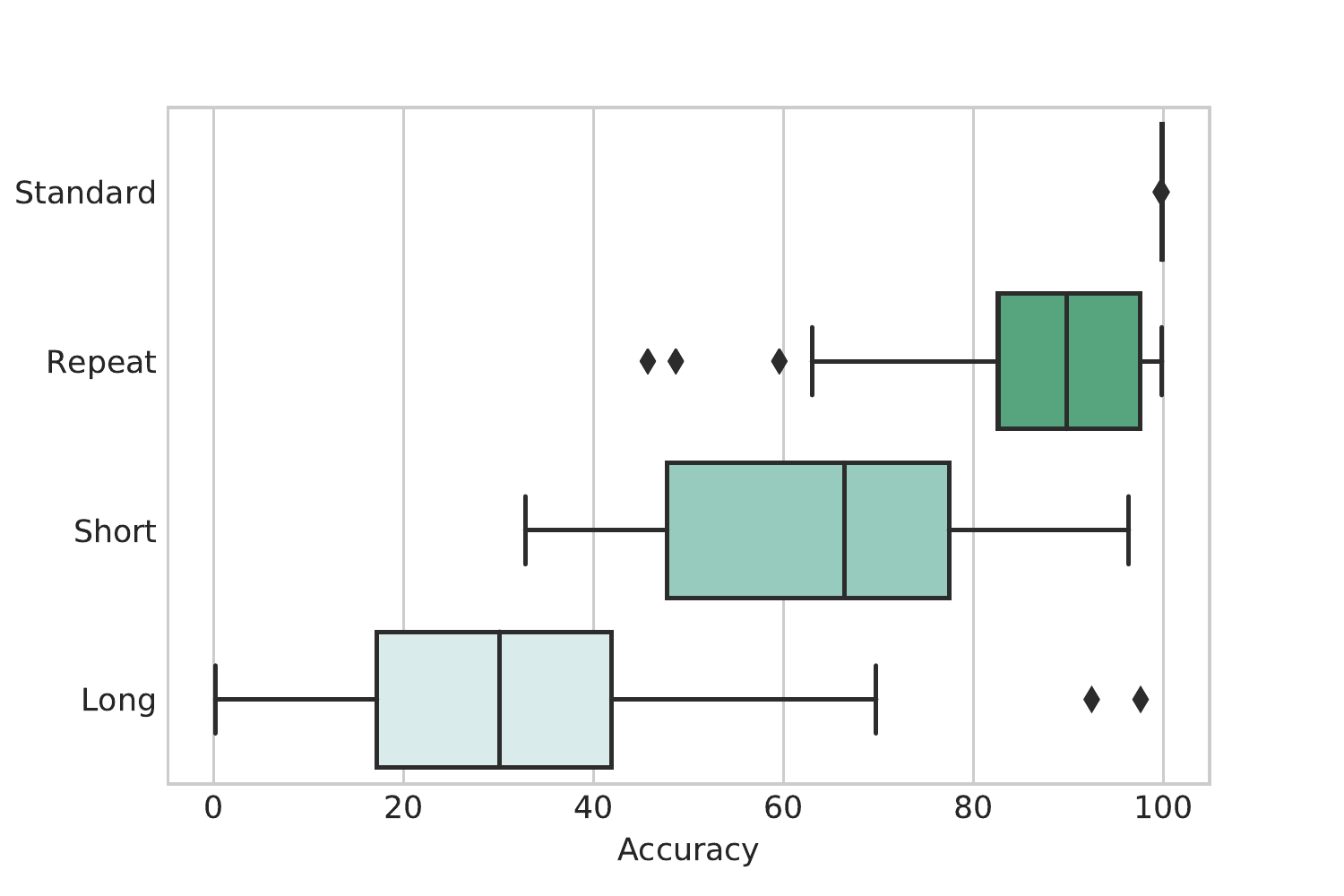}
        \caption{Accuracy \% distribution across 50 runs with different random seeds on the four test sets.}
        \label{fig:teststats-boxplots}
\end{figure}

% For evaluation we trained 50 models with the \textit{same} configuration, chosen with a validation set, with each model trained with a \textit{different} random seed.
For evaluation we trained 50 different models with the \textit{same} configuration, chosen with a validation set, but with \textit{different} random seeds. We created 4 different test sets, each with 2000 randomly generated samples. The first test set consists of samples that are characteristic of the training set, having lengths 5-10 and no repeats (\textbf{Standard}). The second set tests the 
model's ability to generalize to repeated symbols in the input (\textbf{Repeat}). The third and fourth sets test its ability to generalize to different input lengths, strings of length 1-4 (\textbf{Short}) and 11-15 (\textbf{Long}) respectively. Table~\ref{tab:teststats-table} provides further information on the 4 different test sets.
% For evaluation we trained 50 different models with the \textit{same} configuration, chosen with a validation set, but with \textit{different} random seeds. We created 4 different test sets, each with 2000 randomly generated samples. The first test set consists of samples that are characteristic of the training set (\textbf{Standard}). The second set tests the 
% model's ability to generalize to repeated symbols in the input (\textbf{Repeat}). The third and fourth sets test its ability to generalize to different lengths, with input strings of length 1-4 (\textbf{Short}) and 11-15 (\textbf{Long}) respectively.

% this table has characteristic of various test set
\begin{table}%[b]
\centering
\small
\begin{tabular}{l | r | r | r| r }
 & Standard & Repeat & Short & Long  \\
\hline
Size & 2k & 2k & 2k & 2k \\
\hline
Src Length & 5-10 & 5-10 & 1-4 & 11-15 \\     
\hline
Tgt Length & 15-30 & 15-30 & 3-12 & 33-45 \\  
\end{tabular}
\caption{Details about the four test sets used in our experiments.}
\label{tab:teststats-table}
\end{table}

\section{Results}

The distribution of model accuracy measured at instance level on the four test sets across all the 50 seeds is given in Figure~\ref{fig:teststats-boxplots}. All models perform above 99\% on the standard set, with a deviation well below 0.1. However, the deviation on the other two sets is much larger, ranging from 13.39 for the repeat set to 20.63 for the long set. In general, the model performs better on the repeat set than on the short and long sets. Performance on the short and long sets is not always bad, some seeds giving performances of above 95\% for either the short or long set. Ideally, we would like a seed which performs good on all the test sets; however, this seems hard to obtain. The highest average performance across the non standard test sets for any seed was 79.52\%. Learning to generalize for both the repeated and longer inputs seems even harder, with the Pearson correlation between performance on the repeat and long sets being -0.71. 

%However, there is always at least one seed for each test set that achieves an accuracy of more than 95\% on that set. Each test set also had at least one seed that received a score below 50\% on the set (lowest being on the long set at 0.15\%).
%thus
We provide the summary statistics across all 50 runs (50 different random seeds) in Table~\ref{tab:testset-acc-summ}, which gives 
the mean, standard deviation, minimum, and maximum accuracies across all random seeds.  
We additionally provide a sample of performances for some individual random seeds in Table~\ref{tab:testset-acc}, with the highest and lowest accuracies in each column highlighted.

% result for 50 runs: random seeds
\begin{table}[t!]
\centering
\small
\begin{tabular}{r | r | r | r | r}
Seed & Standard & Repeat & Short & Long\\
\hline
2787 & \textbf{99.88} & 94.65 & 42.05 & 23.05 \\
\hline
5740 & 99.86 & \textbf{45.70} & 56.55 & \textbf{97.60} \\
\hline
10000 & 99.86 & 98.55 & \textbf{32.80} & \textbf{0.15} \\
\hline
14932 & \textbf{99.73} & 87.05 & 42.20 & 29.75 \\
\hline
28897 & 99.87 & \textbf{99.85} & 47.40 & 1.40 \\
\hline
30468 & 99.87 & 86.35 & \textbf{96.35} & 12.90 \\
\end{tabular}
\caption{Accuracy \% on the test sets for selected runs out of 50 with different random seeds.}
\label{tab:testset-acc}
\end{table}

% this table has summary of runs with random seeds
\begin{table}[t!]
\centering
\small
\begin{tabular}{l | r | r | r| r }
 & Standard & Repeat & Short & Long  \\
\hline
Mean & 99.85 & 86.67 & 64.36 & 32.09 \\
\hline
Std. & 0.03 & 13.39 & 18.61 & 20.63 \\
 \hline
Min. & 99.73 & 45.70 & 32.80 & 0.15 \\
\hline
Max. & 99.88 & 99.85 & 96.35 & 97.60 \\        
\end{tabular}
\caption{Accuracy \% summarized across all 50 runs with different random seeds.}
\label{tab:testset-acc-summ}
\end{table}

\section{Related Work}
\citet{elman1991distributed} provides one of the earliest works investigating the ability of RNNs to capture properties of language necessary for generalization. Further work explored how RNNs can 
learn context free languages \cite{wiles1995learning,rodriguez1998recurrent} as well some context sensitive languages \cite{schmidhuber2001}. Recent work has moved 
outside the formal language space, with experiments that indicate RNNs may capture the hierarchical structure of natural language syntax, despite not having any hierarchical bias built in the model or the data \cite{linzen2018poverty, Gulordava2018}.

Though these models appear to capture systematic structure necessary 
for generalization, how they do so is often extremely counter intuitive \cite{bowman2018,bernardy2017using}. \citet{lake2018still} has questioned the systematicity with which these models learn. The same work also makes similar 
observations about the difficulty of generalizing to longer length strings. Despite the difficulty models have generalized to longer length strings; they note that the model only needs to see a relatively small amount of longer length strings during training in order to generalize (up to the length of the longest string shown in training).

Our experiments indicate that the final model's ability to generalize 
is highly dependent on the random seed. In should be noted however that the random seed has an effect on several components of the training process, most notably, the exact initialization of the network 
and the order in which training samples are shown. Though the exact value of the initialization likely plays a factor, recent work by  \citet{liska2018} gives evidence that the order 
training data is presented may play an equally large part in determining whether a 
model achieves generalization. 
\section{Conclusions}

The variability in generalization on uncharacteristic inputs (and thus, the extent of linguistic generalization) given different random seeds is alarming, particularly given the fact that the standard test set performance remains mostly the same regardless. The task presented here was easy and simple to analyze, however, future work may be done on natural language tasks. If these properties hold it might indicate that a new evaluation paradigm for NLP should be pushed; one that emphasizes performance on uncharacteristic (but structurally sound) inputs in addition to the data typically seen in training.

\section*{Acknowledgements}
This work is supported in part by the National Science Foundation
under Grant IIS-1617969.

\bibliographystyle{acl_natbib}
\bibliography{naaclhlt2018}
%\appendix
%\newpage
%\section{Supplemental Material}
%\label{sec:supplemental}

\end{document}